\documentclass[sigconf]{acmart}
\AtBeginDocument{%
  }

\copyrightyear{2025}
\acmYear{2025}
\setcopyright{acmlicensed}
\acmConference[MM '25] {Proceedings of the 33rd ACM International Conference on Multimedia}{October 27--31, 2025}{Dublin, Ireland.}
\acmBooktitle{Proceedings of the 33rd ACM International Conference on Multimedia (MM '25), October 27--31, 2025, Dublin, Ireland}
\acmISBN{979-8-4007-2035-2/2025/10}
\acmDOI{10.1145/3746027.3758277}

\usepackage{multirow,diagbox,makecell,bm,bbding}
\usepackage{amsmath,amsthm}
\usepackage{todonotes}
\usepackage{graphicx}
\usepackage{pifont}
\usepackage{algorithm}
\usepackage{algorithmic}
\usepackage{amsfonts}       % blackboard math symbols
\usepackage{booktabs}       % professional-quality tables
\usepackage{footmisc}
\usepackage{tcolorbox}
\usepackage{subcaption}
\usepackage{placeins}
\usepackage{hyperref}

\begin{document}

\title{GEMeX-RMCoT: An Enhanced Med-VQA Dataset for Region-Aware Multimodal Chain-of-Thought Reasoning}

\author{Bo Liu}
\affiliation{%
    % \department{Department of Data Science and Artificial Intelligence,}
  \institution{The Hong Kong Polytechnic University,}
  % \streetaddress{1 Th{\o}rv{\"a}ld Circle}
  \city{Hong Kong S.A.R.}
  \country{China}}
\email{bokelvin.liu@connect.polyu.hk}

\author{Xiangyu Zhao}
\affiliation{%
    % \department{Department of Data Science and Artificial Intelligence,}
  \institution{The Hong Kong Polytechnic University,}
  % \streetaddress{1 Th{\o}rv{\"a}ld Circle}
  \city{Hong Kong S.A.R.}
  \country{China}}
\email{xiang-yu.zhao@connect.polyu.hk}

\author{Along He}
\affiliation{%
    % \department{National Engineering Laboratory for Big Data System Computing Technology,}
  \institution{Shenzhen University,}
  % \streetaddress{1 Th{\o}rv{\"a}ld Circle}
  \city{Shenzhen}
  \country{China}}
\email{healong@szu.edu.cn}

\author{Yidi Chen}
\affiliation{%
    % \department{Department of Radiology,}
  \institution{West China Hospital of Sichuan University,}
  % \streetaddress{1 Th{\o}rv{\"a}ld Circle}
  \city{Chengdu}
  \country{China}}
  \email{chenyidi1152@126.com}

\author{Huazhu Fu}
\authornotemark[1]
\affiliation{%
    % \department{Institute of High Performance Computing,}
  \institution{IHPC, Agency for Science, Technology and Research,}
  % \streetaddress{1 Th{\o}rv{\"a}ld Circle}
  \country{Singapore}}
\email{hzf@ieee.org}

\author{Xiao-Ming Wu}
\authornote{Corresponding authors.}
\affiliation{%
    % \department{Department of Data Science and Artificial Intelligence,}
  \institution{The Hong Kong Polytechnic University,}
  % \streetaddress{1 Th{\o}rv{\"a}ld Circle}
  \city{Hong Kong S.A.R.}
  \country{China}}
\email{xiao-ming.wu@polyu.edu.hk}

\renewcommand{\shortauthors}{Bo Liu, et al.}
\newcommand{\dataset}{\text{GEMeX-RMCoT}}
\begin{abstract}
Medical visual question answering aims to support clinical decision-making by enabling models to answer natural language questions based on medical images. 
While recent advances in multi-modal learning have significantly improved performance, current methods still suffer from limited answer reliability and poor interpretability, impairing the ability of clinicians and patients to understand and trust model outputs. 
To address these limitations, this work first proposes a \textbf{R}egion-Aware \textbf{M}ultimodal \textbf{C}hain-\textbf{o}f-\textbf{T}hought (RMCoT) dataset, in which the process of producing an answer is preceded by a sequence of intermediate reasoning steps that explicitly ground relevant visual regions of the medical image, thereby providing fine-grained explainability.
Furthermore, we introduce a novel verifiable reward mechanism for reinforcement learning to guide post-training, improving the alignment between the model's reasoning process and its final answer.
Remarkably, our method achieves comparable performance using only one-eighth of the training data, demonstrating the efficiency and effectiveness of the proposal. The dataset is available at \url{https://www.med-vqa.com/GEMeX/}.
\end{abstract}





\maketitle

\section{Introduction}
\begin{figure}[!t]
    \centering
    \includegraphics[width=1.0\linewidth]{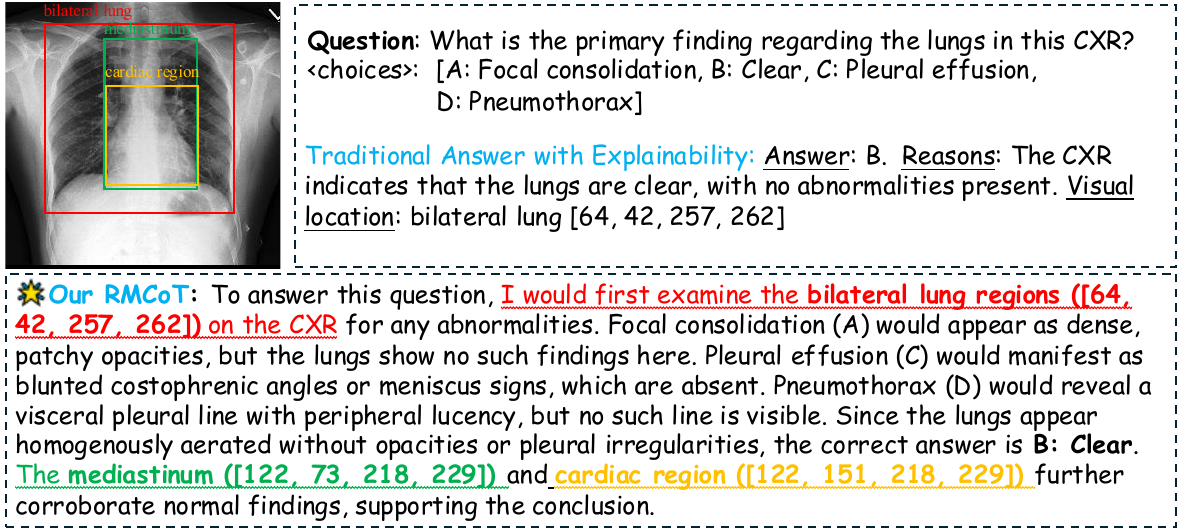}
    \caption{
    (Top) Traditional answer with explainability provided by GEMeX: separate textual and visual prompts; 
    \\($\star$Bottom) Our multimodal, more detailed thinking process for answer generation, which explicitly grounds evidence in specific regions (shown in the colored texts) of the medical image, i.e., anatomical areas that support diagnosis, thereby enhancing the understanding of questions and answers.
    }
    \label{fig:motivation}
\end{figure}

Medical Visual Question Answering (Med-VQA) has emerged as a promising paradigm for supporting clinical decision-making by enabling models to answer natural language questions based on medical images~\cite{lau2018dataset}. Recent advances in multi-modal learning have led to significant performance improvements across a range of Med-VQA benchmarks~\cite{liu2021contrastive, khare2021mmbert, hu2024omnimedvqa, he2020pathvqa}, especially in the era of large vision language models~\cite{li2024llava, zhang2023biomedclip, chen2024eyegpt, sellergren2025medgemmatechnicalreport, cui2025hila,zou2025uncertainty}. However, despite these gains, existing methods remain limited in their ability to provide interpretable and trustworthy responses—an essential requirement for real-world deployment in clinical settings. In particular, most current approaches generate answers without revealing the underlying reasoning process or the specific image regions that inform their decisions. This lack of transparency undermines user trust and hinders the integration of such systems into clinical workflows.

Recently, GEMeX~\cite{liu2024gemex} introduced a new dataset where each VQA triplet is paired with a textual reason and a corresponding visual region as an explanation, bridging this gap to some extent. However, the textual and visual prompts are independent and lack coherence. Moreover, the textual reason merely serves as an explanation of the answer rather than providing guidance on how to solve the question based on the image.
Therefore, based on it, we propose a new dataset \textbf{GEMeX-RMCoT} that enhances interpretability through introducing a \textbf{R}egion-Aware \textbf{M}ultimodal \textbf{C}hain-\textbf{o}f-\textbf{T}hought reasoning mechanism, as shown in Figure~\ref{fig:motivation}. Different from the traditional CoT~\cite{zhang2023multimodal} and recent approaches~\cite{pan2025medvlm,lai2025med} that rely solely on reinforcement learning to incentivize the thinking process, 
our RMCoT explicitly decomposes answer generation into intermediate steps, along with grounding corresponding visual evidence within the medical image. By linking reasoning steps to spatially localized regions~\cite{su2025thinking}, we offer clinicians and patients a fine-grained explanation of how the model arrives at its conclusions.

While the GEMeX-RMCoT dataset enables the modeling of explicit thinking, our empirical findings indicate supervised fine-tuning (SFT) alone is insufficient to fully unleash the reasoning capabilities of the large vision language models (LVLMs). 
To address this, we employ reinforcement learning for post-training that builds upon the SFT-tuned model. It introduces a novel verifiable reward mechanism to incentivize alignment between the reasoning trajectory and relevant visual and textual cues, thereby enhancing both the accuracy of intermediate thinking steps and final answer generation.
In contrast to existing methods requiring large-scale datasets, our approach is data-efficient, achieving comparable performance using only one-eighth of the full training data.

In summary, our contributions are threefold:
\begin{itemize}
    \item We introduce \dataset{}, a region-aware multimodal chain-of-thought dataset for Med-VQA, which provides step-by-step reasoning explicitly grounded in specific anatomical regions of medical images, offering better interpretability for understanding the questions and answers.
    \item We perform supervised fine-tuning on the LVLM using our \dataset{} dataset, and further apply reinforcement learning with a novel verifiable reward mechanism to incentivize reasoning abilities, enabling the generation of more accurate thinking paths and final answers.
    \item  From comprehensive experiments, we demonstrate the effectiveness of the proposed dataset and reward mechanism that empower LVLM to achieve comparable performance with only one-eighth of the training data.
\end{itemize}

\section{Related Work}
\subsection{Medical VQA Dataset}
Recent years have witnessed significant progress in medical visual question answering through the development of specialized datasets targeting diverse clinical challenges. VQA-RAD~\cite{lau2018dataset} established the foundation with 3,000 question-answer pairs focused on radiology images. SLAKE~\cite{liu2021slake} expanded the scope with over 14,000 manually annotated QA pairs spanning CT, MRI, and X-ray modalities. OmniMedVQA~\cite{hu2024omnimedvqa} further broadened coverage across multiple body regions and imaging modalities to enhance model generalization capabilities. For specialized tasks, PathVQA~\cite{he2020pathvqa} offers 32,000 QAs on histopathology images. MIMIC-Diff-VQA~\cite{hu2023expert} emphasizes differential diagnosis between paired X-rays. GEMeX~\cite{liu2024gemex} enhances VQA with both textual and visual explanations, facilitating a better understanding of the answers. However, none of them incorporates a reasoning process for problem-solving, which hinders patients and junior doctors from fully understanding the questions and answers. To fill this gap, we propose a region-aware multimodal CoT reasoning dataset, \dataset{}, in this work.

% In recent years, various datasets have been developed to advance medical VQA, each addressing specific clinical challenges. VQA-RAD~\cite{lau2018dataset} pioneered the field with 3,000 QA pairs on radiology images. SLAKE~\cite{liu2021slake} introduced over 14,000 manually annotated QAs across CT, MRI, and X-ray modalities. OmniMedVQA~\cite{hu2024omnimedvqa} broadens coverage across body regions and modalities to promote generalization. 

% PathVQA~\cite{he2020pathvqa} offers 32,000 QAs on histopathology images. For specialized tasks, RadGenome-Chest CT~\cite{zhang2024radgenome} focuses on chest CT diagnostics; MIMIC-Diff-VQA~\cite{hu2023expert} emphasizes differential diagnosis between paired X-rays; and MIMIC-CXR-VQA~\cite{bae2024ehrxqa} augments MIMIC-CXR~\cite{johnson2019mimic} with diverse thoracic QAs. 
% However, existing datasets lack rich explanations and varied question formats, limiting their utility for explainable reasoning and real-world applicability.

\subsection{Reinforcement Learning}
Recently, the growing popularity of reasoning-focused large language models such as GPT-o1~\cite{jaech2024openai} and DeepSeek-R1~\cite{guo2025deepseek} has drawn increasing attention to the underlying mechanisms of reinforcement learning (RL). This interest has led to remarkable advancements in tasks such as mathematics~\cite{luong2024reft,ying2024internlm}, code generation~\cite{jiao2024preference, zhang2024codedpo}, hallucination detection~\cite{zhao2023beyond, sun2023aligning}, and interdisciplinary research \cite{zhao2025msearth, zhou2025scientists}. A notable contribution is the Group Relative Policy Optimization (GRPO) algorithm proposed in DeepSeekMath~\cite{shao2024deepseekmath}, which significantly simplifies the training process. Unlike traditional RL algorithms like PPO~\cite{jiao2024preference} that require a critic model to evaluate policy performance, GRPO compares groups of candidate responses directly, eliminating the need for an additional critic.
GRPO’s simplicity has inspired further exploration into the reasoning capabilities of LVLMs for multimodal and visual tasks~\cite{huang2025vision, liu2025visual, su2025openthinkimg}. In the medical domain, GRPO has also been applied to incentivize the reasoning abilities of LVLMs for tasks such as out-of-distribution detection~\cite{pan2025medvlm} and visual question answering~\cite{lai2025med}. Unlike these works that merely elicit textual descriptions of visual content, we propose a visually grounded reasoning paradigm that enhances interpretability and strengthens evidential support.

% \section{Construction of \dataset{}: Thinking with Visual Grounding in Medical VQA}
\section{Construction of \dataset{}}

\begin{figure*}[!t]
    \centering
    \includegraphics[width=1.0\linewidth]{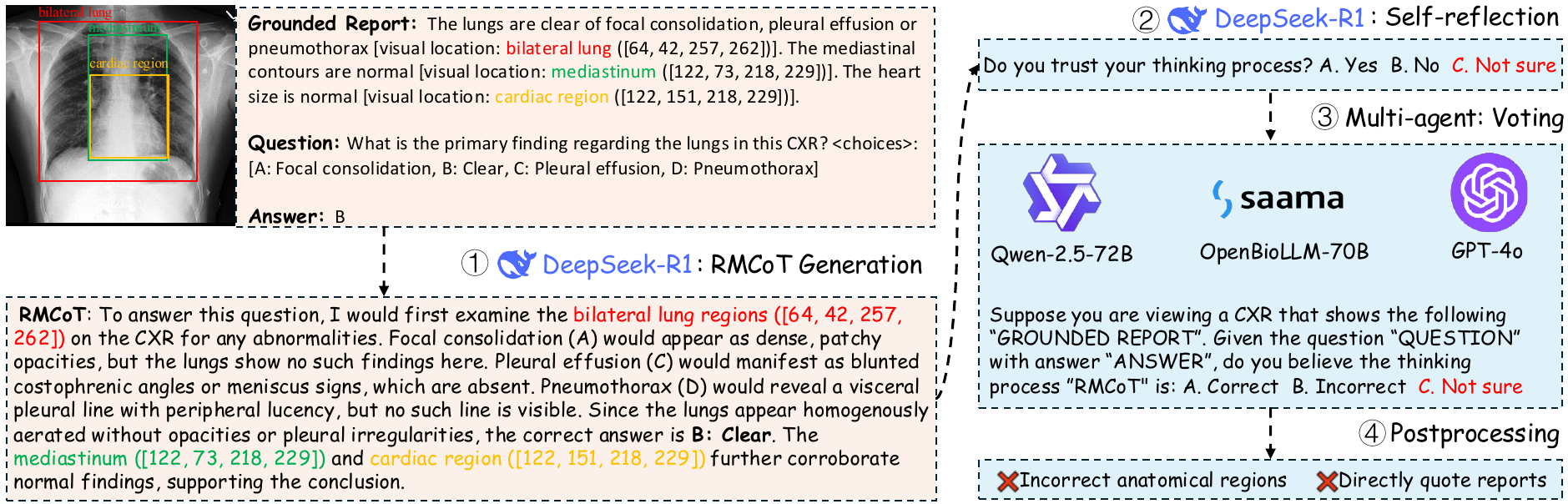}
    \caption{
   Illustration of the proposed uncertainty-driven multi-agent framework for generating the RMCoT dataset. The left (orange) section depicts the data generation process, while the right (blue) section shows the uncertainty-driven quality assurance pipeline, i.e., the ``not sure'' option prompts a deeper reflection and more deliberate decision-making.
    }
    \label{fig:framework}
\end{figure*} 

\subsection{Task Definition} 
Unlike conventional chain-of-thought approaches used in general tasks such as VQA~\cite{wang2025multimodal}, which typically generate only textual rationales before outputting the final answer, we propose a multimodal-level thinking process with visual grounding tailored for Med-VQA. As illustrated in Figure~\ref{fig:motivation}, it decomposes the reasoning process into several steps with visual attention to specific image regions, aiming to deliver more detailed explanations and improved interpretability for patients and junior doctors.
Formally, given a question $q$ for image $v$, we hope to provide a multi-modal thinking process $T$ involving visual grounding to attain answer $a$ via
\begin{equation}
T = \left( \{r_i\}_{i=1}^{N_r}, \{t_j\}_{j=1}^{N_t} \right),
\end{equation}
where $\{r_i\}_{i=1}^{N_r}$ is a set of image regions and $\{t_j\}_{j=1}^{N_t}$ is a set of relevant textual descriptions.

\subsection{An Uncertainty-Driven Multi-Agent Framework}
As illustrated in Figure~\ref{fig:framework}, we construct the RMCoT dataset based on GEMeX~\cite{liu2024gemex}, the largest Med-VQA dataset that provides detailed visual and textual explanations for each VQA triplet. To ensure the high quality of RMCoT, we introduce an uncertainty-driven multi-agent framework, centered around DeepSeek-R1~\cite{guo2025deepseek}, which also coordinates multiple agents and leverages an uncertainty mechanism to enhance the accuracy of the generated data.

\subsubsection{Data Generation} In Stage \textcircled{1}, we primarily use DeepSeek-R1 to generate the thinking process for each VQA triplet due to its outstanding reasoning capabilities. 
However, as DeepSeek-R1 is a text-based LLM and lacks the ability to directly perceive visual inputs, we utilize the grounded reports from GEMeX as an intermediary. In GEMeX's reports, each sentence is aligned with a specific anatomical region, enabling DeepSeek-R1 to simulate a multimodal reasoning process required for solving the tasks.
To optimize the prompt design, we randomly sample 50 questions from each question type (e.g., open-ended) in the training set—resulting in 200 questions in total—to evaluate performance and guide prompt iteration. The final optimized prompt is presented in Table~\ref{tab:prompt_cot}.

\begin{table}[t!]\centering
\begin{minipage}{1.0\columnwidth}\vspace{0mm}    \centering
\begin{tcolorbox} 
    \centering
      \footnotesize
\begin{tabular}{p{0.95\columnwidth} c}
{\bf messages} = [``role'':``{\bf user}'', ``content'': \\
f```Suppose you are viewing a CXR that shows the following: ``The hilar contours are normal [visual location: bilateral hilar structures ([116, 112, 227, 182])] ...''. Given the question: ``YOUR QUESTION'', provide a detailed thinking process (around 100 words) including specific visual location (e.g., (region [x1,y1,x2,y2])) about how to solve this question with answer ``ANSWER''. You must assume that you are viewing the CXR image rather than reading the textual findings, thus do not output words like ``observe the report'' or ``from report'' or ``report states'' or ``given findings'' or ``provided findings'' or ``described findings''.""
\end{tabular}
\end{tcolorbox}
% \vspace{-2mm}
\caption{Given a question-answer pair, our tailored prompt guides DeepSeek-R1 to generate RMCoT data.}

\label{tab:prompt_cot}
\end{minipage}
\end{table}

\subsubsection{Uncertainty-driven Quality Assurance} Although manual verification and prompt modification are applied, the generated data still suffers from quality issues. To address this, we design a pipeline which contains three steps (Stage \textcircled{2} to Stage \textcircled{4}) to ensure quality: 
(1) We first introduce an uncertainty-aware mechanism that enables DeepSeek-R1 to self-reflect and evaluate the reliability of its own outputs. Specifically, after generating the RMCoT response, we prompt DeepSeek-R1 with the question: \emph{``Do you trust your thinking process? A. Yes B. No C. Not sure''}. This self-reflection with uncertainty choice ``Not sure'' allows the model to carefully reassess its previous reasoning, which has been shown to improve output accuracy~\cite{chen2023quantifying}. We retain only those RMCoT responses for which the model answers “A. Yes.”;
(2) To avoid bias caused by relying on a single LLM for both data generation and evaluation~\cite{panickssery2024llm}, we introduce additional agents to assess the quality of the generated data. As shown in Stage \textcircled{3}, we employ three additional agents—Qwen-2.5-72B~\cite{qwen2024qwen25}, OpenBioLLM-70B\footnote{https://huggingface.co/aaditya/Llama3-OpenBioLLM-70B}, and GPT-4o~\cite{achiam2023gpt}—as external reviewers. These agents are given grounded reports, questions, and answers to assess the accuracy of the thinking process. Similar to the self-reflection stage, we introduce an ``uncertain'' option to provide the agents with greater flexibility for consideration and decision-making. This process helps eliminate incorrect reasoning paths generated by DeepSeek-R1 based on its internal knowledge. Then, we adopt a voting mechanism: the RMCoT is retained only if all three agents unanimously select ``A. Correct'' option; 
(3) Finally, we apply post-processing to eliminate RMCoT outputs with significant errors, such as incorrect location information or instances that quote the report or findings directly, rather than generating content from an observational perspective.

\subsection{Radiologist Validation}
After auto-generation, a radiologist randomly samples 200 examples from each of the four question types in the dataset for manual inspection. Cross-validation reveals no clinically inconsistent reasoning chains, demonstrating the effectiveness of the proposed framework. In total, we generate 202,384 RMCoT instances for 21,994 images, all drawn from the GEMeX training set. The detailed distribution is presented in Table~\ref{tab:statistics}.

\begin{figure}[!t]
    \centering
    \includegraphics[width=0.98\linewidth]{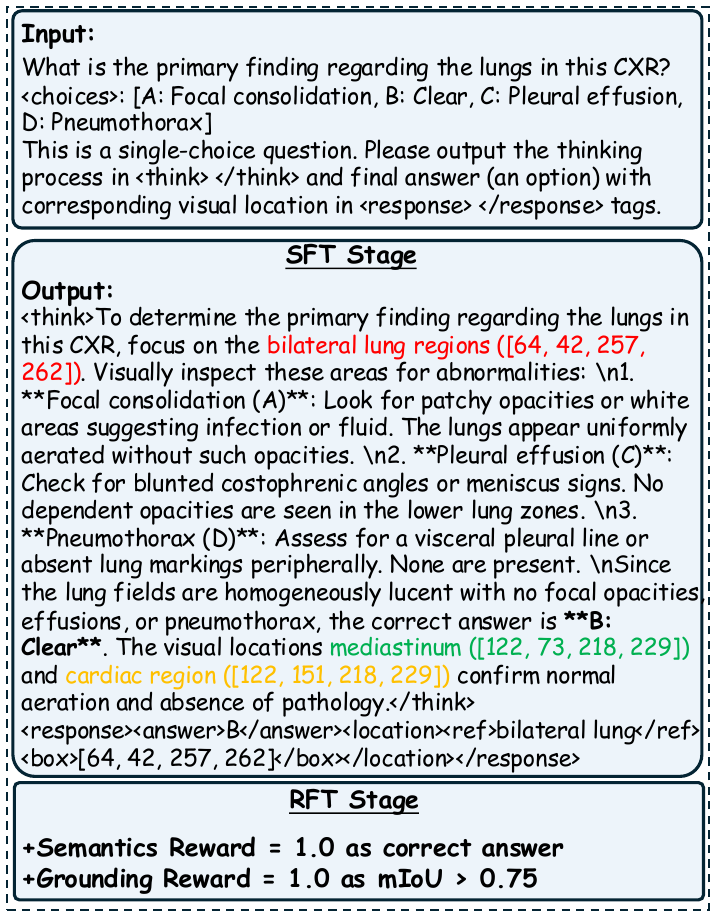}
    \caption{
    An example from the SFT and RFT stages: after SFT, the model learns to think carefully and incorporates grounding to specific pathological regions for visual evidence. During the RFT stage, since both the answer and the grounded regions are correct, the two reward scores are each 1.0.
    }
    \label{fig:example}
\end{figure}

\begin{table}[!t]
    \centering
    \resizebox{1\linewidth}{!}{
    \begin{tabular}{cccc}
    \toprule
    Open-ended  & Closed-ended  & Single-choice  & Multi-choice \\
    \hline
    61,240  & 36,432  & 52,850 & 51,862 \\
    \bottomrule
    \end{tabular}}
    \caption{Distribution of generated RMCoT data across the four question types in GEMeX.} 
    \label{tab:statistics}
\end{table}

\section{Fine-tuning on \dataset{}: Learning to Think with Visual Grounding in Med-VQA}

\subsection{Supervised Fine-tuning}

After generating the RMCoT data, we first train the large vision language model (LVLM) using Supervised Fine-tuning (SFT) to equip it with task-aware thinking capabilities. 
To generate the outputs required by GEMeX—which include both the final answer and its corresponding visual location (omitting the reasons, as we provide a more detailed preceding thinking path)—the model must produce structured outputs in two distinct parts. First, it should output a thinking trace that incorporates explicit visual grounding, enclosed within the \texttt{<think>}...\texttt{</think>} tags. Second, it should deliver a concise final response that includes the predicted ``answer'' and the summarized ``visual location'', wrapped within a designated XML-like structure: \texttt{<response>} \texttt{<answer>}...\texttt{</answer>} \texttt{<location>}...\texttt{</location>}\texttt{</response>}. This format ensures both interpretability and alignment with the GEMeX evaluation requirements.
An illustrative example is shown in Figure~\ref{fig:example}.

\subsection{Reinforcement Fine-tuning}
Although SFT can significantly improve thinking capabilities, we have observed in practice that: (1) due to the complexity of the task, the model sometimes generates incorrect thinking process after SFT, for example, incorrect analysis of images, which may be because SFT is better at helping LVLMs memorize rather than truly understand ~\cite{chu2025sft}; (2) inaccurate region localization within the thinking path can propagate errors to the final answer, ultimately leading to degraded performance.
Therefore, we adopt Reinforcement Fine-tuning (RFT), specifically through Group Relative Policy Optimization (GRPO)~\cite{shao2024deepseekmath}, as a post-training strategy to further activate and enhance the model’s thinking capabilities with a newly proposed reward mechanism.

\subsubsection{Group Relative Policy Optimization} Given a question $q$ from question set $Q$, GRPO first samples a set of outputs $\{o_i\}_i^G$ from the policy model $\pi_{\theta_{old}}$ during each learning iteration. Then it optimizes the LVLM $\pi_{\theta}$ by maximizing the following objective:
\begin{equation}
\label{eq:grpo}
{\scalebox{0.85}{$
\begin{aligned}
\quad \quad \mathcal{J}_{\text{GRPO}}(\theta) &= \mathbb{E}_{q \sim P(Q),\, \{o_i\}_{i=1}^G \sim \pi_{\theta_{\text{old}}}(O|q)} \Bigg[ 
 \\ 
\frac{1}{G} \sum_{i=1}^G  
 \min \Bigg(  
\frac{\pi_{\theta}(o_i \mid q)}{\pi_{\theta_{\text{old}}}(o_i \mid q)} A_i, 
 &
 \text{clip} \left(
\frac{\pi_{\theta}(o_i \mid q)}{\pi_{\theta_{\text{old}}}(o_i \mid q)}, 1 - \varepsilon, 1 + \varepsilon
\right) A_i \Bigg) \\
&  - \beta D_{\text{KL}}(\pi_{\theta} \parallel \pi_{\text{ref}}) \Bigg]
\end{aligned}
$}}
\end{equation}
where $A_i = \frac{r_i - \text{mean}(\{r_1, r_2, \ldots, r_G\})}{\text{std}(\{r_1, r_2, \ldots, r_G\})}$ is the relative advantage over group rewards $\{r_i\}_i^G$, $D_{\text{KL}}$ is the KL divergence between $\pi_{\theta}$ and $\pi_{\text{ref}}$ (the model after SFT stage) to prevent catastrophic forgetting caused by over optimization,  $\epsilon$ and $\beta$ are the PPO clipping hyper-parameter~\cite{schulman2017proximal}
and the coefficient for the KL penalty, respectively.

\begin{table}[t!]\centering
\begin{minipage}{1.0\columnwidth}\vspace{0mm}    \centering
\begin{tcolorbox} 
    \centering
      \footnotesize
\begin{tabular}{p{0.98\columnwidth} c}
{\bf messages} = [``role'':``{\bf user}'', ``content'': 
f```We would like to request your feedback on the performance of two AI assistants in response to the user question displayed above. For your reference, the visual content in the image is represented with caption describing the same image.
Please rate the accuracy (most important), relevance of their responses, considering both answer and reason (if any). Each assistant receives an overall score on a scale of 1 to 10, where a higher score indicates better overall performance.
Please output both scores and your reaon in JSON format \{ assistant1: score, assistant2: score, reason:your\_reason \}."""
\end{tabular}
\end{tcolorbox}
% \vspace{-2mm}
\caption{Our designed prompt for \emph{Semantics Reward}.}
\label{tab:prompt_reward}
\end{minipage}
\end{table}

\begin{table*}[!t]
    \centering
    \small
    \renewcommand{\arraystretch}{1.3}
\resizebox{1\linewidth}{!}{
    \begin{tabular}{c|ccc|cc|cc|cc|cc|c}
    \toprule
    \hline
    %  \hline
      &  \multicolumn{3}{c|}{Training Paradigm} &  \multicolumn{2}{c|}{Open-ended}&\multicolumn{2}{c|}{Closed-ended} & \multicolumn{2}{c|}{Single-choice}& \multicolumn{2}{c|}{Multi-choice} & \\
     
     &  Dataset & SFT & RFT &  A-score\dag & V-score  & A-score & V-score& A-score  & V-score & A-score &  V-score & A-score.Avg \\
     \hline
     Zero-shot & \multicolumn{3}{c|}{-} & 85.13 & 12.31 & 30.39 & 20.23 & 49.00 & 27.81 & 12.02 &  7.59 & 44.14\\ 
    
     \hline
     \multirow{4}{*}{Fine-tuning} & GEMeX-Full & \Checkmark & - & \textbf{97.31} & \textbf{60.51} & 82.32 & \textbf{69.32} & \underline{83.08} & \textbf{67.38} &  \textbf{74.51} & \textbf{55.98} & \underline{84.31}\\ 
     &GEMeX-200K  & \Checkmark  & -  & 92.54 &56.59 & 79.93 &  66.04 & 78.92 & 61.39 & 67.93 & 49.71 & 79.83\\
     &\dataset{}-200K & \Checkmark & \XSolidBrush & 94.71 & 56.52 & \underline{87.11} & 66.23 & 80.08 & 61.76 & 70.20 & 51.13 & 83.03\\  
     & \dataset{}-200K & \Checkmark & \Checkmark & \underline{96.44} & \underline{59.63} & \textbf{90.24}	&	\underline{68.39}	& \textbf{83.31}	 & 	\underline{64.52} & \underline{73.18} & \underline{54.49} & \textbf{85.79} \\ 
     \hline
    \bottomrule
    \end{tabular}
    }
    \caption{Performance of Qwen2.5-VL-7B with different evaluating paradigms on GEMeX. The A-score indicates answer or choice accuracy (\%), and the V-score represents mIoU (\%). $\dag$ indicates that the accuracy of the answer is judged by the LLM, as the question is open-ended. The best results are bolded, and the second-best are underlined in each column.} 
    \label{tab:vqa_result}
\end{table*}
\subsubsection{Reward Functions}
We propose two reward functions, tailored for the two output parts of GEMeX: answer and involved visual locations. (1) \emph{Semantics Reward}: In GEMeX, four types of questions are defined (Table~\ref{tab:statistics}). 
For choice-based and closed-ended questions, reward design is straightforward, as we can directly use Accuracy Reward~\cite{pan2025medvlm} to evaluate whether the model’s outputs are correct.
Nevertheless, designing a reward function for open-ended questions is particularly challenging due to the absence of fixed answers. 
Unlike~\cite{liang2025group} which employs traditional machine learning metrics such as BLEU as rewards—metrics that have been shown to poorly capture essential medical content~\cite{jain2021radgraph}—we introduce a semantics-aware accuracy reward. It enables the integration of all four types into a unified, coherent reward paradigm.
Specifically, we input both the ground truth (GT) answer and the model-generated answer into OpenBioLLM-70B, which can provide two accuracy scores at the semantics level. When these two scores are close, i.e., if the absolute difference between them is less than 2, we assign a reward of 1 point; otherwise, the reward is 0. The specific prompt is shown in Table~\ref{tab:prompt_reward}. (2) \emph{Grounding Reward}: Besides giving answers, the model needs to provide the visual locations involved in solving the questions. Therefore, we first check whether the number of bounding boxes predicted by the model matches the number in the GT. If they match, we then evaluate whether the mean Intersection over Union (mIoU) exceeds 0.75. A reward score of 1.0 is assigned only when both conditions are satisfied; otherwise, the reward is 0.
% Note that we leave out the common \emph{format reward} as after the SFT stage, the model has learned to follow the format requirement.

\section{Experiments}
\subsection{Dataset}
We conduct experiments on the GEMeX dataset and randomly sample a portion of its training set to generate RMCoT data. We denote the version with the complete training set of 1.59M VQA triples as GEMeX-Full. The 200K subset that includes RMCoT data is denoted as \dataset{}-200K, while the subset without RMCoT is referred to as GEMeX-200K. For evaluation, we use the original GEMeX test set, which consists of 300 images and 3,960 questions (1,144 open-ended questions, 543 closed-ended questions, 1,300 single-choice questions, and 973 multiple-choice questions).

\subsection{Training Details}
We mainly explore the Qwen2.5-VL-7B-Instruct for experimental verification. During the SFT stage, we fine-tune both the visual projection layers and the LLM components. Particularly, the model is trained for 2 epochs on 8 NVIDIA H100 GPUs with a batch size of 256. The network is warmed up in the ﬁrst 0.05 epochs with a linear learning rate from 3e-7 to 1e-4, which further decays by cosine schedule. The optimizer is AdamW. 
During the RFT stage, the model is trained for 1 epoch on 8 NVIDIA H100 GPUs with a batch size of 128. The learning rate is $2e-6$. Regarding the hyper-parameters in Eq.~\ref{eq:grpo}, we generate $8$ rollouts (i.e., $G$) for each input and set $\beta$ and $\epsilon$ to $1e-3$ and $0.2$, respectively. The sample generator and LLM used for semantics reward are deployed by vLLM~\cite{kwon2023efficient}.

\subsection{Main Results}
We report the main results in Table~\ref{tab:vqa_result}. It can be seen that (1) Compared to zero-shot learning, fine-tuning can significantly improve the model's performance;
(2) When SFT is conducted with only 200K vanilla data (i.e., GEMeX-200K), the model lags behind the one trained on the full dataset by a large margin, with an average A-score that is around 4.5\% lower; 
(3) When RMCoT data (GEMeX-RMCoT-200K) is further used for SFT, the model's performance improves substantially, even surpassing that of the model trained on the full dataset in certain tasks, e.g., an improvement of approximately \textbf{5\%} on closed-ended tasks. Nevertheless, it can be observed that when only SFT is used, the improvement in visual grounding is quite limited. This is often due to inaccurate location reasoning in the thinking path after learning RMCoT with SFT, which leads to error propagation in the final outputs;
(4) To address this issue, further applying RFT with semantics reward and grounding reward yields noticeable enhancement, a \textbf{1.5\%–3\%} improvement in both answer accuracy and localization performance. It is worth noting that, compared to using the full training data, the model further fine-tuned with RFT achieves better results in answer generation (i.e., higher average A-score) and comparable performance in localization. These findings demonstrate the effectiveness of our proposed RMCoT dataset and RFT rewards.

\begin{table}[!t]
    \centering
    \renewcommand{\arraystretch}{1.2}
    \resizebox{1.0\linewidth}{!}{
    \begin{tabular}{ccc|cccc}
    \toprule
    Task & SFT Data &Original A-score & Per.1 & Per.2 & Per.3 & Std.\dag \\
    \hline
    % \multirow{2}{*}{Closed.}  & 200K  & 79.93  & 76.98 & 83.06 & 75.69 &  78.58 $\pm$ 3.94\\
    \multirow{2}{*}{Closed.}  & 200K  & 79.93  & 76.98 & 83.06 & 75.69 &  3.214\\
      % & RMCoT-200K  & 87.11  & 88.39 &  86.56 & 87.66 & 87.54 $\pm$ 0.92\\
      % \hline
    & RMCoT-200K  & 87.11  & 88.39 &  86.56 & 87.66 & \textbf{0.752}\\
      \hline
    % \multirow{2}{*}{Single.}  &  200K  & 78.92  & 79.00 & 77.69 & 77.31 & 78.00 $\pm$ 0.89\\ %0.723
        \multirow{2}{*}{Single.}  &  200K  & 78.92  & 79.00 & 77.69 & 77.31 & 0.723\\ %
      % &  RMCoT-200K  & 80.08  & 80.15 &  80.00 & 79.69 & 79.95 $\pm$ 0.23 \\ %0.192
            &  RMCoT-200K  & 80.08  & 80.15 &  80.00 & 79.69 & \textbf{0.192} \\ %
    \hline
    \bottomrule
    \end{tabular}}
    \caption{Model performance under different perturbations. $\dag$ means the standard deviation across the three perturbations from original A-score.} 
    \label{tab:robustness}
\end{table}

\begin{figure*}[!t]
    \centering
    \includegraphics[width=0.93\linewidth]{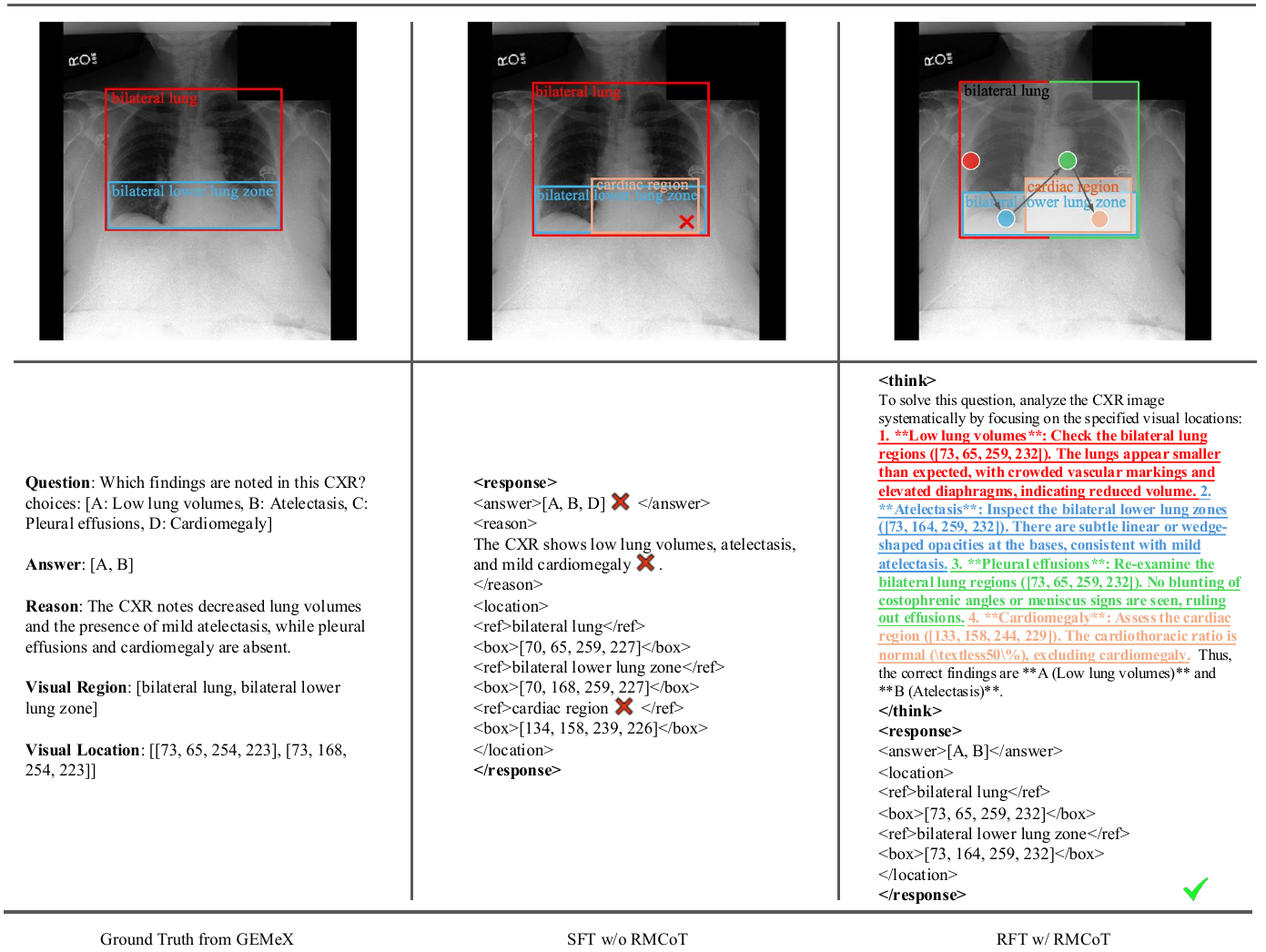}
    \caption{
One challenging example from GEMeX answered by models trained with and without RMCoT. (\Checkmark) or (\XSolidBrush) in outputs highlight correct or incorrect reasons or answers. The colored words indicate the thinking with visual grounding process.
    }
    \label{fig:evaluation_2}
\end{figure*}

\subsection{Robustness to Input Variations}
We conduct two common perturbations to generate input variations: (1) we modify closed-ended questions while preserving their semantic integrity—for example, changing ``Is the heart size abnormal in this CXR?'' by randomly replacing the interrogative word ``is'' with ``isn't'', or replacing ``abnormal'' with ``normal'', and adjusting the corresponding answer accordingly. During the test phase, each question has a 50\% chance of being modified by one of the two modes mentioned above. If a question does not meet the criteria for either mode, no modification will be applied; (2) we randomly change the order of the options in single-choice questions. For each type of perturbation, we conduct three rounds of test and show models' performance in Table~\ref{tab:robustness}. To more clearly compare the effectiveness of RMCoT, we compare the performance of the models only after SFT. The results show that after fine-tuning the model with RMCoT, its robustness against interference is significantly improved (i.e., lower standard deviation), highlighting another advantage, especially in the context of potential challenges encountered in practical deployment.

\subsection{Qualitative Evaluation}
As shown in Figure~\ref{fig:evaluation_2}, we compare the performance of models trained with and without RMCoT on GEMeX-200K. When using RMCoT, we present the model outputs after RFT stage. It can be seen that when using RMCoT with proper tuning, not only are the answers more accurate, but the model also carefully thinks with analyzing corresponding regions in the image when addressing questions, leading to stronger overall interpretability. More examples are shown on the project page.

% \begin{figure*}[!t]
%     \centering
%     \includegraphics[width=0.93\linewidth]{imgs/show_cases_1.pdf}\\
%     \includegraphics[width=0.93\linewidth]{imgs/show_cases_2.pdf}
%     \caption{Challenging examples from GEMeX answered by models trained with and without RMCoT. (\Checkmark) or (\XSolidBrush) in outputs highlight correct or incorrect reasons or answers. The colored words indicate the thinking with visual grounding process.}
%     \label{fig:evaluation_1}
% \end{figure*}

\section{Conclusion}
In this work, we address the critical issue of interpretability in Medical Visual Question Answering by introducing a region-aware multimodal chain-of-thought dataset (RMCoT). It explicitly links thinking steps to corresponding regions in the medical image, offering fine-grained visual explanations that enhance transparency and trust in model-generated answers. To further guide the model in generating accurate thinking paths, we proposed a new reward mechanism for reinforcement learning that aligns the model’s outputs with relevant visual and textual evidence. 
Despite using a subset of training data, the trained model is able to achieve impressive performance.
We believe this work marks a step toward making Med-VQA systems more explainable and clinically usable, paving the way for safer and more reliable AI-assisted diagnosis.

\clearpage
\begin{acks}
We thank all reviewers for their valuable time and feedback. This work was supported by Project P0056021 under Parent Project P0050643 of the Otto Poon Charitable Foundation Smart Cities Research Institute, HK PolyU; the Research Student Attachment Programme, HK PolyU; and the A*STAR Central Research Fund (``Robust and Trustworthy AI system for Multi-modality Healthcare''), Singapore.
\end{acks}

\bibliographystyle{ACM-Reference-Format}
\balance
\bibliography{sample-base}

\end{document}